\documentclass[conference]{IEEEtran}
\usepackage{mathtools}

\ifCLASSINFOpdf
\else
\fi
\renewcommand\IEEEkeywordsname{Keywords}

\begin{document}
%
\title{Predictive Coding as Stimulus Avoidance\\ 
in Spiking Neural Networks}

\author{\IEEEauthorblockN{Atsushi Masumori}
\IEEEauthorblockA{Graduate School of Arts and Sciences\\
The University of Tokyo\\
Tokyo, Japan\\
masumori@sacral.c.u-tokyo.ac.jp}
\and
\IEEEauthorblockN{Takashi Ikegami}
\IEEEauthorblockA{Graduate School of Arts and Sciences\\
The University of Tokyo\\
Tokyo, Japan\\
ikeg@sacral.c.u-tokyo.ac.jp\\}
\and
\IEEEauthorblockN{Lana Sinapayen}
\IEEEauthorblockA{Sony Computer Science Laboratories, Inc.\\
Tokyo, Japan\\
}
\IEEEauthorblockA{
Earth-Life Science Institute\\
Tokyo, Japan\\
lana.sinapayen@gmail.com}
}



%


\maketitle

\begin{abstract}
Predictive coding can be regarded as a function which reduces the error between an input signal and a top-down prediction. If reducing the error is equivalent to reducing the influence of stimuli from the environment, predictive coding can be regarded as stimulation avoidance by prediction. Our previous studies showed that action and selection for stimulation avoidance emerge in spiking neural networks through spike-timing dependent plasticity (STDP). In this study, we demonstrate that spiking neural networks with random structure spontaneously learn to predict temporal sequences of stimuli based solely on STDP.
\end{abstract}


%
\IEEEpeerreviewmaketitle

\section{Introduction}
Prediction has recently been argued to be a important function of the brain \cite{Bastos2012, Clark2013} and predictive coding \cite{Rao1999, Huang2011} has attracted the attentions of researchers in many fields \cite{Friston2009, Lotter2017}. Predictive coding can be regarded as a function for reducing errors between an input signal and a top-down prediction. If reducing errors is equivalent to reducing the influence of environmental stimuli, predictive coding can be regarded as stimulation avoidance by prediction. 
Our previous studies showed that spiking neural networks with spike-timing dependent plasticity (STDP) \cite{Song2000} learn to avoid stimuli from the environment by their action \cite{Sinapayen2016, Masumori2017a}. Cultured neural networks learn actions to avoid stimulation in the same way \cite{Masumori2017b}. In addition, we found that neuronal cultures avoid stimulation by selecting what external information is received or declined \cite{Masumori2018}. 
In our previous study, we demonstrated that spiking neural networks with asymmetric STDP, in which the dynamics of long-term potentiation and long-term depression are rotational asymmetric, reproduced such selection. 
Therefore, based on STDP, two types of stimulation avoidance emerge: stimulation avoidance by action and stimulation avoidance by selection.


In this study, we evaluate whether prediction also emerges in spiking neural networks based on STDP. Several studies have examined the prediction of temporal stimulation in spiking neural networks \cite{Buonomano2000, Rao2001, Wacongne2012}. However, these predictive networks require specifically designed network topology or other synaptic functions than STDP, such as short-term plasticity. We hypothesise that there is no need to include such structures or functions other than STDP for learning to predict a simple sequence of stimuli because our previous studies showed that neural networks with random initial weights learn to avoid stimuli by their action based on STDP, and for hidden neurons, a prediction that inhibits the input neurons response to stimulation is equivalent to an action that eliminates the stimulation.



We first demonstrate that minimal predictive networks consisting of 3 to 6 neurons with synaptic weight governed by STDP spontaneously learn to predict sequences of stimuli. We then show that even larger random networks (of 100 neurons) without a specifically designed structure spontaneously learn to predict sequences of stimuli based solely on STDP. 

\section{Methods}
\subsection{Izhikevich Neuron Model}
The spiking neuron model proposed by Izhikevich \cite{Izhikevich2003a} was used to simulate excitatory and inhibitory neurons consisting small and large networks. This model is widely applied as indivisual parameters can be adjusted to reproduce the dynamics of many types of neurons, and it is also computationally efficient. The basic equations of this neural model are as follows: 

\begin{eqnarray}
\begin{split}
&\frac{dv}{dt} = 0.04v^2 + 5v + 140 -u + I,\\
&\frac{du}{dt} = a(bv - u),\\
&\text{if } v \geq 30 \text{~mV},~\text{then}
\begin{cases}
    v \leftarrow c\\
    u \leftarrow u+d.
\end{cases}
\end{split}
\end{eqnarray}

\begin{figure}[htbp]
\centering
\includegraphics[width=9cm]{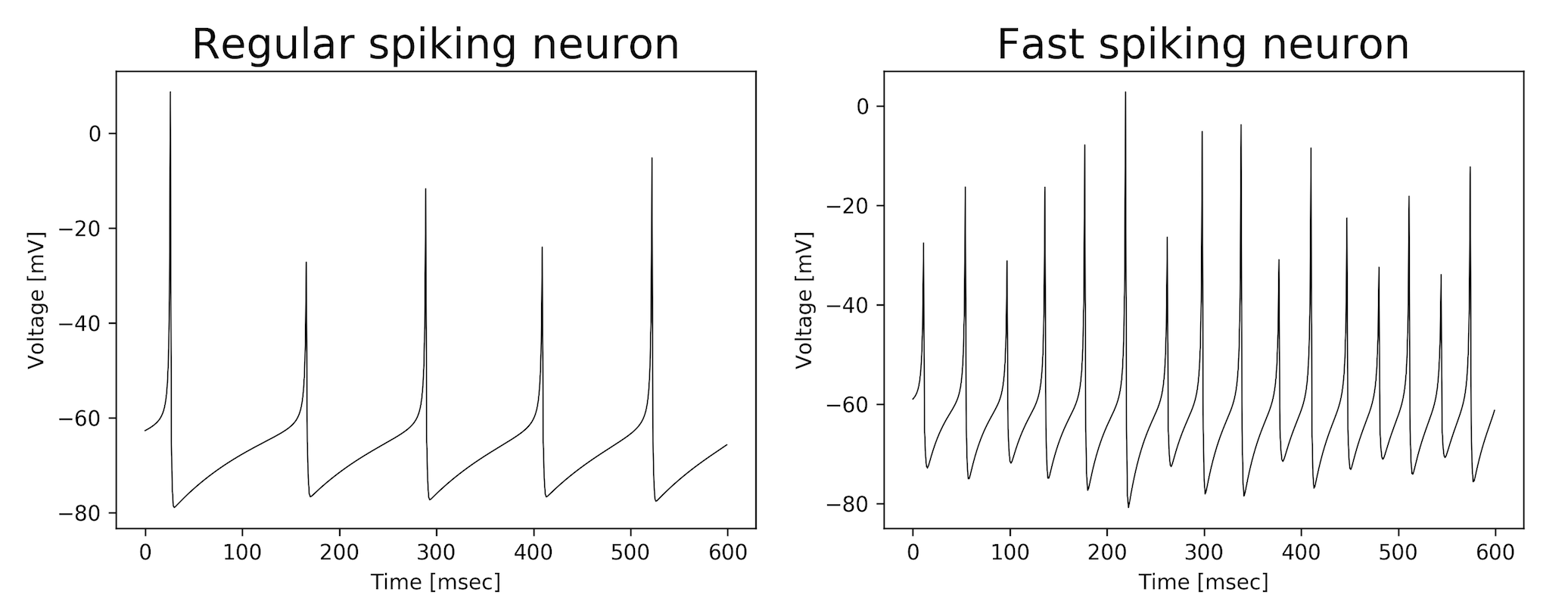}
\caption[Firing patterns of regular-spiking and fast-spiking simulated using the Izhikevich model.]{Firing patterns of regular spiking and fast spiking neurons simulated using the Izhikevich model. Regular-spiking neurons are used as excitatory neurons and fast-spiking neurons are used as inhibitory neurons.}
\label{fig:IzhDynamics}
\end{figure}

Here, $v$ represents the membrane potential of the neuron, $u$ is a variable related to membrane repolarization, $I$ represents the input current (with multiple components as explained below), $t$ is time, and $a, b, c$, and $d$ are parameters controlling the shape of the spike \cite{Izhikevich2003a}. The neuron is regarded as firing when the membrane potential $v \geq 30$ mV. The parameters for excitatory neurons were set to $a = 0.02,~b = 0.2,~c = -65\text{~mV}$, and $~d = 8$, and the parameters for inhibitory neurons to $a = 0.1,~b = 0.2,~c = -65\text{~mV}$, and $~d = 2$. With these parameters, excitatory neurons show regular spiking and inhibitory neurons show fast spiking (Fig.~\ref{fig:IzhDynamics}). The simulation time step $\Delta t$ was set to 1~ms.

The variable $I$ represents depolarization evoked by synaptic currents, noise, and external stimuli, and was added to the membrane potential of each neuron $n_i$ at every time step as follows:

\begin{equation}
\label{eq:neuronInput}
\begin{split}
&I_i = \sum_{j=0}^{n}f_j w_{ji} + e_i + m_i\\
&f_j =  
\begin{cases}
    1 , & \text{if neuron $j$ is firing}\\
    0, & \text{otherwise}.
\end{cases}
\end{split}
\end{equation}

Here, $w_{ji}$ represents the weight of individual synapse between presynaptic neuron $j$ to postsynaptic neuron $i$, where weights are positive for synapse from excitatory neurons and negative for synapses from inhibitory neurons, $m$ is zero-mean Gaussian noise with standard deviation $\sigma=3$~mV which represents the internal noise, $e$ represents the external stimulation (with conditions, frequency and strength of external stimulation varying among on experiments).

In most of the experiments, there was no synaptic delay; however, in the experiments with the longer temporal stimulus sequence (described below), a synaptic delay were added between an action potential of the presynaptic neuron and postsynaptic potential. In these experiments, the function $f_j$ (Eq.~\ref{eq:neuronInput}) was modified by the synaptic delay according to

\begin{equation}
\label{eq:neuronInput_modified}
\begin{split}
&f_j =  
\begin{cases}
    1 , & t - ts_j= td_{ij}\\
    0, & \text{otherwise}.
\end{cases}
\end{split}
\end{equation}

where $t$ denotes the current simulation time, $ts_j$ represents spike timing of presynaptic neuron $j$, and $td_{ij}$ represents the synaptic delay between neuron $i$ and neuron $j$. In the experiments with synaptic delay, each pair of excitatory and inhibitory neuron was connected by 15 synapses and the $td$ of each synapse was varied from 1 to 15 ms.  

\subsection{Spike-Timing Dependent Plasticity}
Spike-timing dependent plasticity was used as the mechanism for changing synaptic weights between spiking neurons. In this model, synaptic weight increased when the presynaptic neuron fires before the postsynaptic neuron and decreased when the presynaptic neuron fires after the postsynaptic neuron. The weight variation $\Delta w$ is defined by

\begin{equation}
\label{eq:delta_weight}
\Delta w =  
\begin{cases}
    A(1-\frac{1}{\tau})^{\Delta t} , &\text{if } \Delta t > 0\\
    -A(1-\frac{1}{\tau})^{-\Delta t} , &\text{if } \Delta t < 0.\\
\end{cases}
\end{equation}

\begin{figure}[htbp]
\includegraphics[width=7.65cm]{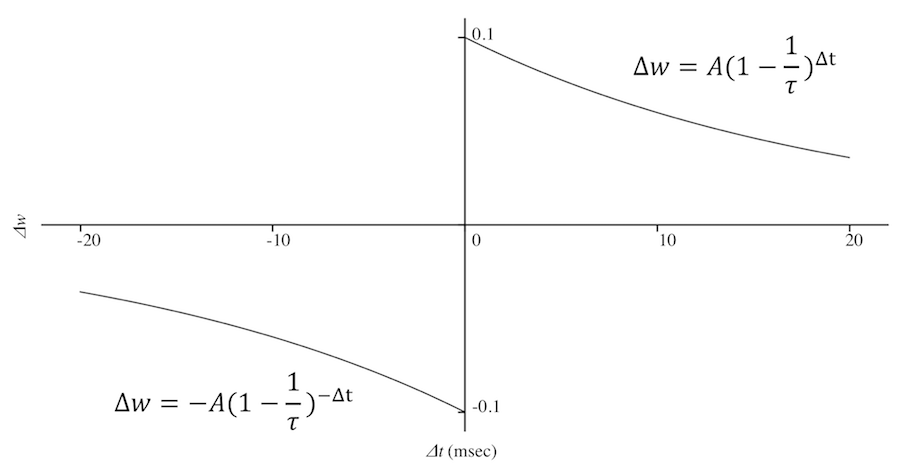}
\caption{The spike-timing depended plasticity (STDP) function used to change the synaptic weights between neurons. The weight variation $\Delta w$ of a synapse from neuron $a$ to neuron $b$ depends on the relative spike timing ($A = 0.1$, $\tau=20$ ms).}
\label{fig:stdp}
\end{figure}

Here, $\Delta t$ represents the relative spike timing between presynaptic neuron $a$ and postsynaptic neuron $b$: $\Delta t = t_b - t_a$ ($t_a$ represents the spike timing of neuron $a$, and $t_b$ represents the spike timing of neuron $b$). For excitatory synapses, $A = 0.1$ and $\tau=20$ ms. Figure~\ref{fig:stdp} shows the variation of $\Delta w$ depending on $\Delta t$; $\Delta w$ is negative when the postsynaptic neuron fires first and positive when the presynaptic neuron fires first. Note that STDP was applied not only to the connections between excitatory neurons but also to connections from inhibitory neurons to excitatory neurons. Although STDP at inhibitory synapse is still controversial, here we applied the reverse shape of the STDP function in Figure~\ref{fig:stdp} for the inhibitory synapses ($A=-0.1$, $\tau=20$).

The weight value $w$ varies as

\begin{eqnarray}
w_t = w_{t-1}+ \Delta w \;.
\label{eq:wvariation}
\end{eqnarray}

The maximum possible weight was fixed at $w_{max}=80$ for excitatory synapses and $w_{max}=0$ for inhibitory synapses, and if $w > w_{max}$,  $w$ was reset to $w_{max}$. Alternatively, the minimum possible weight was fixed at $w_{min}=0$ for excitatory synapses and $w_{min}=-80$ for inhibitory synapses, and if $w < w_{min}$,  $w$ was reset to $w_{min}$. 


 

\subsection{Experimental Setup}
We first performed experiments with minimal predictive networks consisting of 3 to 6 neurons that learn to predict temporal sequences of stimuli. We then evaluated whether this learning performance is scalable to larger random networks of 100 neurons. 

\begin{figure}[htbp]
\begin{center}
\includegraphics[width=9cm]{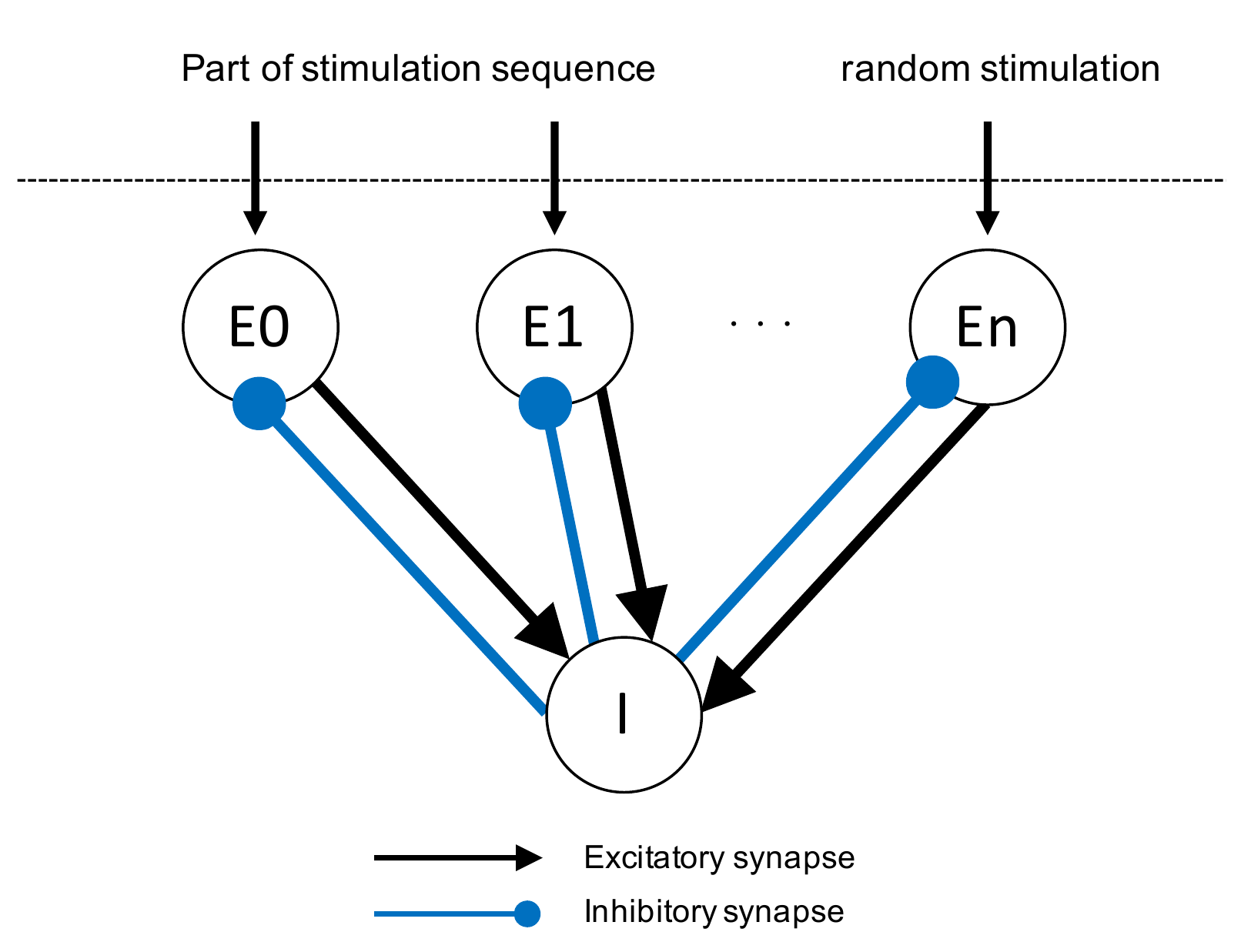}
\end{center}
\caption[Basic topology of a minimal predictive network.]{Basic topology of a minimal predictive network. The network consists of excitatory input neurons and one inhibitory neuron. $E$ represents the excitatory neurons and $I$ represents the single inhibitory neuron. The number of excitatory intput neurons was varied from 3 to 5 across the experiments. Neuron $E_{n}$ receives random stimulus input as control, while others ($E_i$) receive part of the specific stimulation sequence. The excitatory neurons were not connected to each other, but all projected an output to and received an input from inhibitory neuron. The black arrows represent excitatory synapses and the blue line with circles represent inhibitory synapses.}
\label{fig:prediction_network_topology}
\end{figure}

Figure~\ref{fig:prediction_network_topology} shows the basic network topology of the minimal predictive networks used in the first series of experiments, where $E_n$ represents excitatory neurons and $I$ represents an inhibitory neuron. The network consisted of excitatory input neurons and one inhibitory neuron. The input neurons were not connected to each other, but all were connected to and from the inhibitory neuron. The number of input neurons was varied from 3 to 5 across experiments. The initial weight values $w$ between neurons were set to 15 (in arbitrary units). 

We also constructed larger networks to evaluate the scalability of the predictive networks. These larger networks consisted of 80 excitatory neurons and 20 inhibitory neurons, and the network topology was random (i.e., neurons were fully connected with random weights). The weight values $w$ were randomly initialized between 0 and 5 ($0<w<5$) with uniform distributions for excitatory synapse and between -5 and 0 ($-5<w<0$) with uniform distributions for inhibitory synapse. Synaptic plasticity was applied to all connections except for connections between inhibitory neurons. There were three input neuron groups ($EG0$--$EG2$), each consisting of 10 excitatory neurons.

\begin{figure}[htbp]
\begin{center}
\includegraphics[width=9cm]{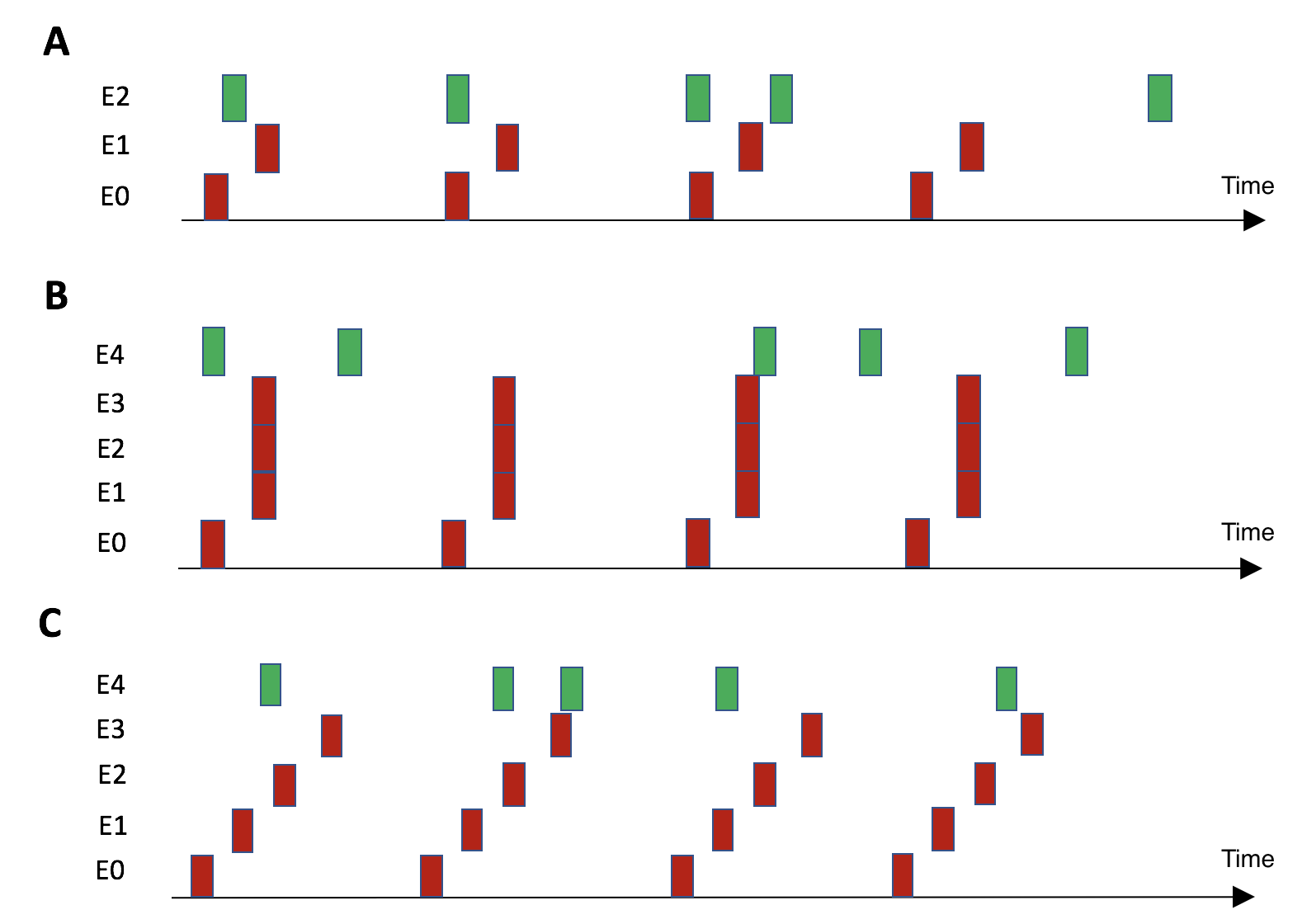}
\end{center}
\caption[Examples of stimulation sequence pattern. A: Minimal pattern. B: Spatially extended pattern. C: Temporally extended pattern.]{Stimulation sequence patterns. The red squares represent the individual applied stimuli within each stimulation sequence, while the green squares represent random stimuli. A: Minimal pattern. The sequence consisted of two stimuli, an initial signal stimulus and subsequent target stimulus. B: Spatial extended pattern. The sequence consisted of four stimuli, one initial signal stimulus and three subsequent target stimuli. C: Temporal extended pattern. The sequence also consisted of four stimuli, one initial signal stimulus and three successive target stimuli.}
\label{fig:stim_seq}
\end{figure}

Three types of stimulus sequences were applied to the networks: a minimal pattern, a spatially extended pattern, and a temporally extended pattern (Fig.~\ref{fig:stim_seq}). The minimal pattern consisted of two stimuli, one signal stimulus followed by one target stimulus, where the signal stimulus was delivered to a specific input neuron and after a fixed time delay the target stimulus was delivered to another specific input neuron (Fig.~\ref{fig:stim_seq}A). Therefore, the timing of the signal stimulus was random (unpredictable) but the timing of the target stimulus was predictable. For the large networks, the minimal pattern consisted of two groups of synchronous stimuli with each stimulus group delivered to the corresponding input neuron group. 

In the spatially extended pattern, the sequence consisted of four stimuli, one signal stimulus and three subsequent target stimuli. The signal stimulus was delivered to a specific input neuron and after a fixed time delay the target stimuli were delivered simultaneously to the three other specific input neurons (Fig.~\ref{fig:stim_seq}B). 

In the temporally extended pattern, the sequence also consisted of four stimuli, one signal stimulus and three subsequent target stimuli. The signal stimulus was delivered to a specific input neuron and after a fixed time delay the first target stimulus was delivered to another specific input neuron followed by successive target stimuli at fixed time interval to other neurons (Fig.~\ref{fig:stim_seq}C). 

For every sequence pattern, each stimulus producess a 100 mV depolarization in the minimal netowrk and a 10 mV depolarization in the larger networks. The duration of each stimulus was set to 1~ms, the interval between each stimulus in the sequence to 10~ms (witch is sufficiently smaller than the 20 ms working time window of STDP), and the interval between each sequence to 300~ms (which is sufficiently larger than the working time window of STDP). In addition to sequential stimulation, random stimulation was delivered into another specific input neuron (or input neuron group in large networks) as the control.

Prediction is defined here as suppression of the input neurons by the inhibitory neuron(s) at the time of stimulus input. This decrease the influence of environmental stimulation on the network; thus, prediction can be regarded as stimulus avoidance.

\section{Results}
\subsection{Minimal Networks}
\begin{figure}[htbp]
\begin{center}
\includegraphics[width=9cm]{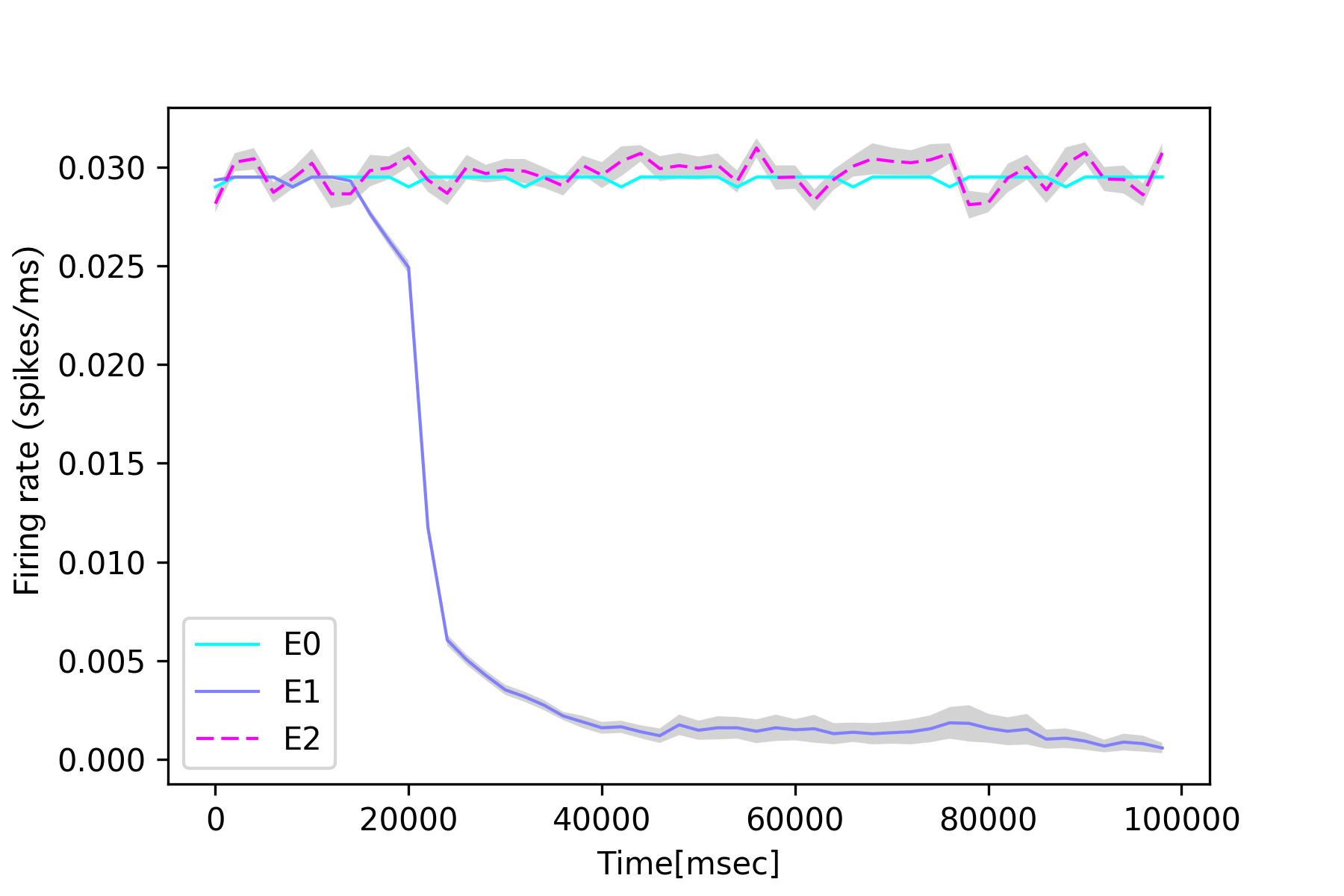}
\end{center}
\caption[Time series of input neuron firing rates within small networks in response to the minimal pattern.]{Time series of input neuron firing rates within small networks in response to the minimal pattern. The shaded regions represents the standard error of the mean (n~=~20 networks). The firing rate of neuron $E1$ receiving the target stimulus decreased substantially while the firing rates of $E0$ receiving the signal stimulus and $E2$ receiving the random stimulus changed little.}
\label{fig:minimal_firing}
\end{figure}

We first examined predictive coding by the smallest minimal network in response to the minimal stimulation pattern without synaptic time delay. The network consisted of three excitatory input neurons and one inhibitory neuron. The input neurons were not connected to each other, but all were connected bidirectionally to the inhibitory neuron (Fig.~\ref{fig:prediction_network_topology}). If the target inputs are correctly predicted, spiking of input neurons should be inhibited at the timing of stimulation. We examined the firing rates of all input neurons and found that the firing rates of neuron $E1$ receiving the target stimulus decreased, whereas the firing rates of neuron $E0$ receiving the signal stimulus and $E2$ receiving the random stimulus did not change substantially (Fig.~\ref{fig:minimal_firing}). Therefore, the network gradually learned to predict the target stimuli (and exclude it) while the random stimulus was not predicted. 

\begin{figure}[htbp]
\begin{center}
\includegraphics[width=8cm]{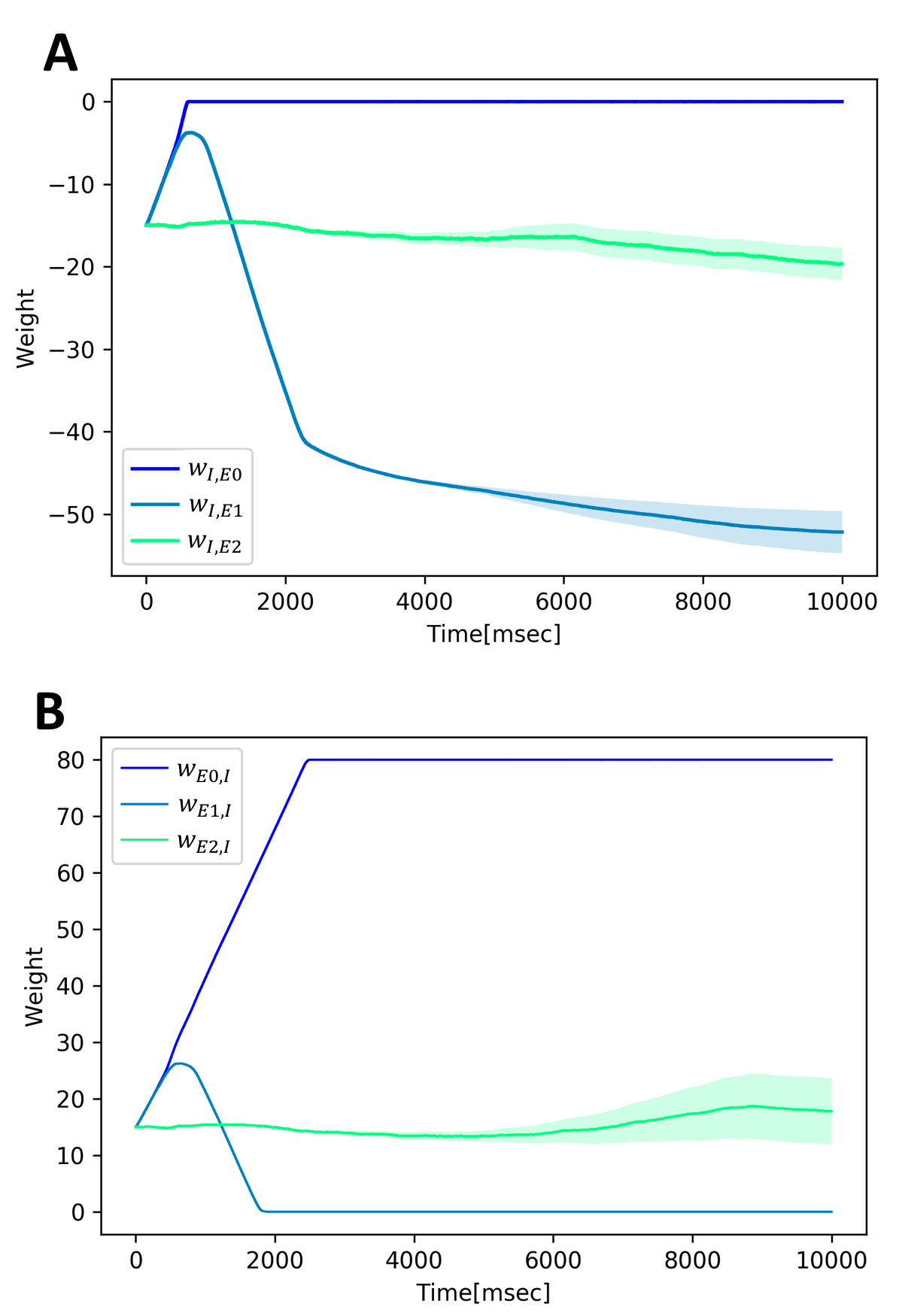}
\end{center}
\caption[Time series of synaptic weight changes in small networks receiving the minimal stimulation pattern.]{Time series of synaptic weight changes in small networks receiving the minimal stimulation pattern. The shaded regions represent the standard error of the mean (n = 20 networks). A: Weights from the inhibitory neuron to the excitatory neurons: $E0$, $E1$, and $E2$. B: Weight from the excitatory neurons: $E0$, $E1$, and $E2$ to the inhibitory neuron. }
\label{fig:minimal_prediction_weight}
\end{figure}

\begin{figure}[htbp]
\begin{center}
\includegraphics[width=8cm]{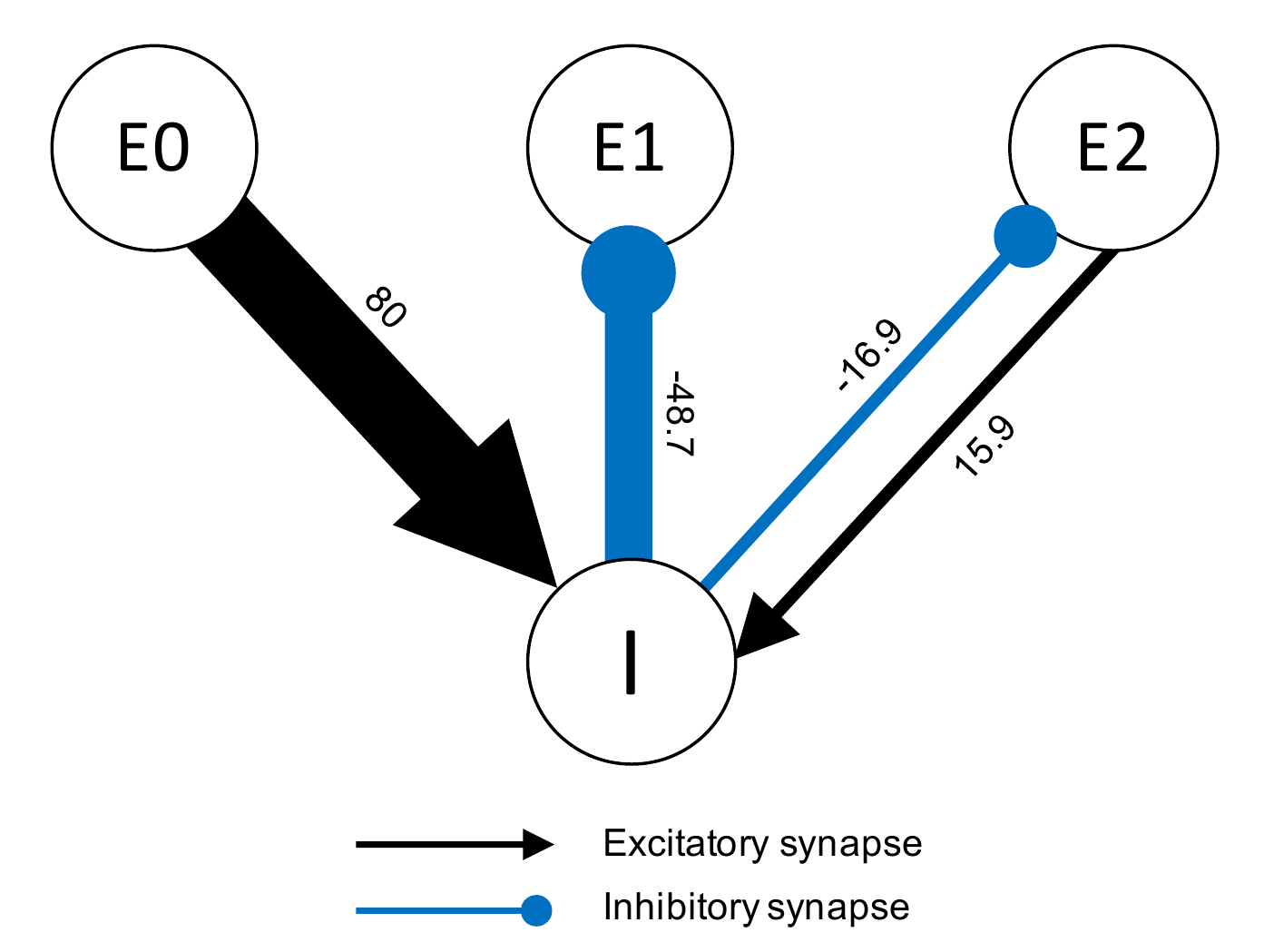}
\end{center}
\caption[Final topology of the smallest network after stimulation with the minimal pattern.]{Final topology of the smallest network after stimulation with the minimal pattern. The pathway from $E0$ to the inhibitory neuron and from the inhibitory neuron to $E1$ was strengthened. This pathway is required to predict target stimuli at $E1$. The black arrows represent the excitatory synapses and the blue connections represents inhibitory synapses. The weight value of connections from the inhibitory neurons is negative.}
\label{fig:prediction_final_topology}
\end{figure}

Figures~\ref{fig:minimal_prediction_weight} and \ref{fig:prediction_final_topology} show that the weight of the synapse from the neuron receiving signal stimulation (the signal neuron $E0$) to the inhibitory neuron increased with time, while the weight of the syanapse from the inhibitory neuron to the target neuron ($E1$) decreased with time. The path from $E0$ to $I$ to $E1$ is required to predict the timing of target stimuli at $E1$. In contrast, the weights to and from the random neuron $E2$ changed little. 

\begin{figure}[htbp]
\begin{center}
\includegraphics[width=9cm]{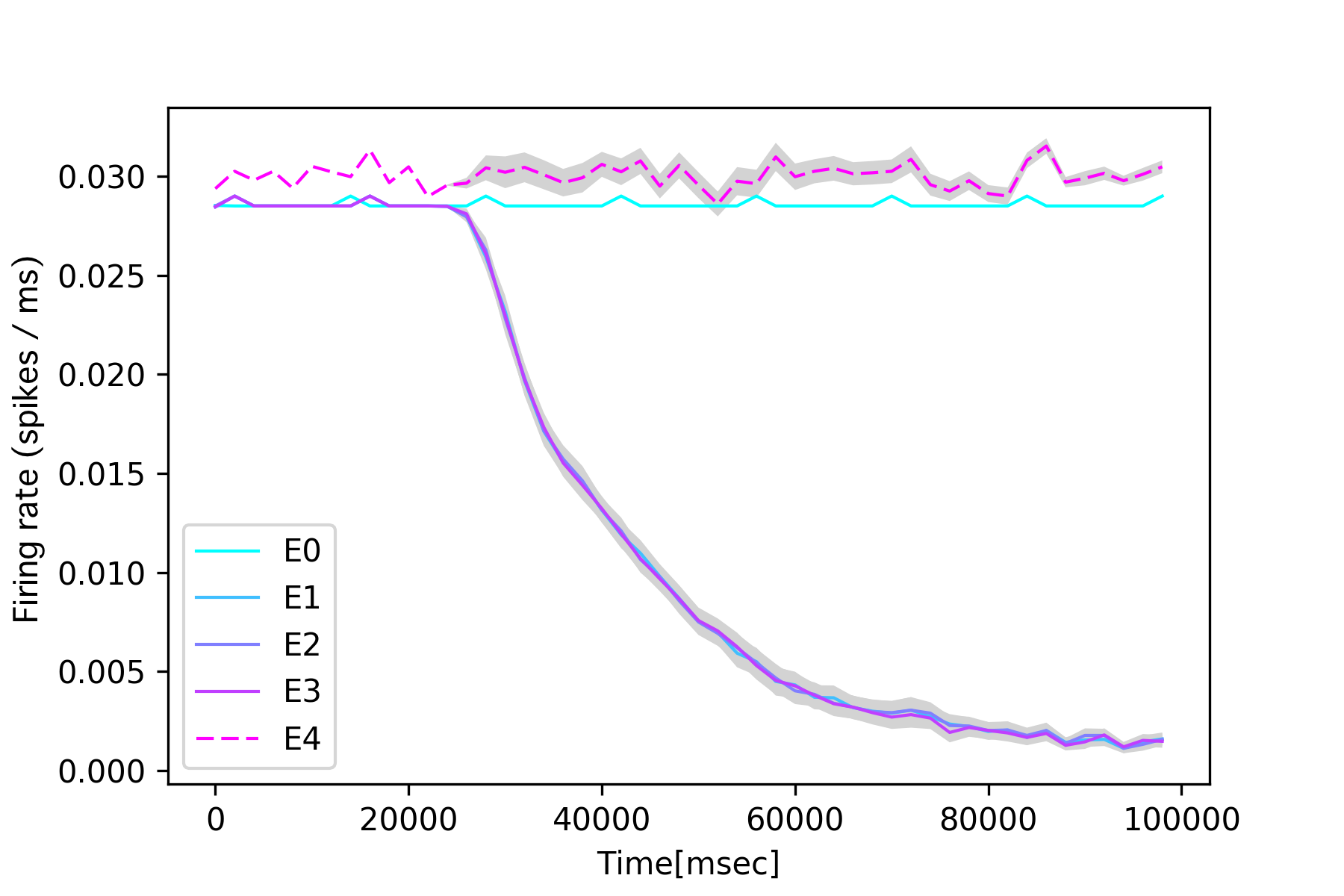}
\end{center}
\caption[Time series of input neuron ($E0$--$E4$) firing rates within small networks in response to the spatially extended stimulation pattern.]{Time series of input neuron ($E0$--$E4$) firing rates within small networks in response to the spatially extended stimulation pattern. The shaded regions represent the standard error of the mean (n = 20 networks). The firing rates of $E1$--$E3$ receiving target stimuli in the spatially extended pattern decreased remarkedly, while $E0$ receiving the signal stimulus and $E4$ receiving the random stimulus changed little.}
\label{fig:spatial_prediction_firing}
\end{figure}

We then evaluated whether small networks could learn to predict more complex spatially extended stimulus patterns (Fig.~\ref{fig:stim_seq}B) and temporally extended stimulus patterns (Fig.~\ref{fig:stim_seq}C). We first applied the spatially extended pattern of stimulation to networks with two more additional input neurons compared to the smallest minimal network shown in Fig.~\ref{fig:prediction_network_topology}. Figure~\ref{fig:spatial_prediction_firing} shows that the firing rates of neurons $E1$ to $3$ receiving target stimuli in the spatial extended pattern decreased with time, whereas the firing rates of neuron $E0$ receiving the signal stimulus and $E4$ receiving the random stimulus were largely unchanged. Thus, these small networks gradually learned to predict the spatially extended target stimulus pattern, while the random stimulus was not predicted. This implies that the inhibitory neuron suppressed the firing of neurons $E1$--$E3$ at the time of target stimulation and that the network learned to predict the spatially extended pattern.

\begin{figure}[htbp]
\begin{center}
\includegraphics[width=9cm]{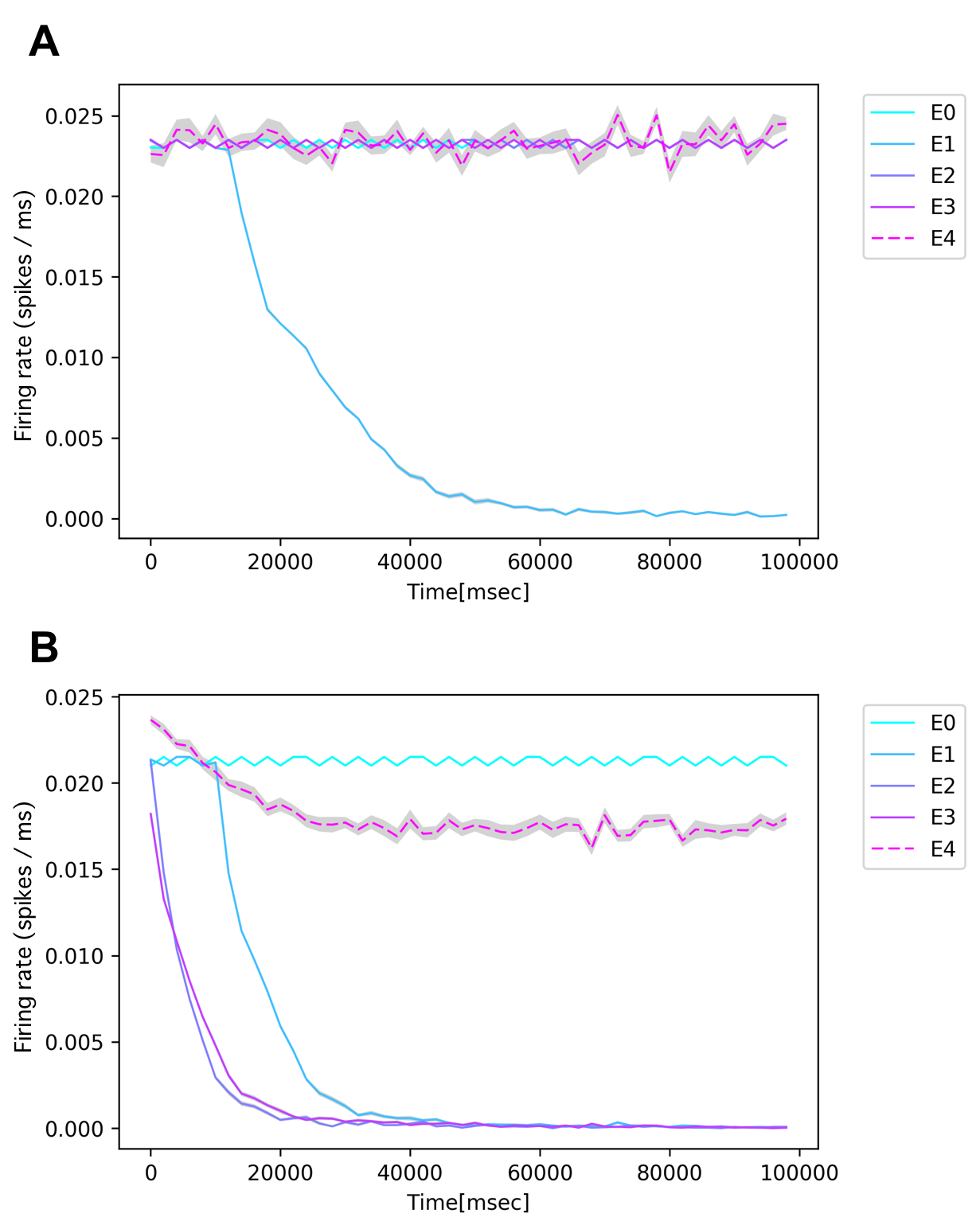}
\end{center}
\caption[Time series of the firing rate of the input neurons with the temporal sequence.]{Time series of the firing rate of the input neurons with the temporal extended pattern. The shaded regions represent standard errors of the mean (n = 20 networks). A: Without synaptic time delay. In this case, only the firing rate of $E1$ which got the first target stimuli decreased. This means that the network without synaptic time delay cannot learn to predict longer temporal sequence. B: With synaptic time delay. In this case, the firing rate of $E1$-$3$ which got the target stimuli of the temporal sequence decreased but $E0$ which got the signal stimulus and $E4$ which got the random stimulus did not change much. This means that the network with synaptic time delay learn to predict longer temporal sequence.}
\label{fig:temporal_prediction_firing}
\end{figure}

On the other hand, these small network did not learn to predict the temporal extended pattern. Figure~\ref{fig:temporal_prediction_firing}A shows that the only firing rate of neuron $E1$ decreased with time. This implies that the network cannot learn to predict the longer temporal sequence than the minimal pattern consists of two stimuli, possibly, because there is only one inhibitory neuron and one connection to each excitatory neuron and longer temporal information cannot be encoded.

For neural networks to predict a longer temporal sequence, we hypothesized that they must have more inhibitory neurons or more synapses between the input neurons and inhibitory neurons with different synaptic delays to encode temporal information. Therefore, we constructed a small model with 15 synapses between each input neuron and the inhibitory neuron and set different time delays (from 1 to 15 ms). 

Figure~\ref{fig:temporal_prediction_firing}B shows that the firing rates of neurons $E1$-$E3$ receiving target stimuli in the temporal sequence decreased whereas the firing rates of $E0$ receiving the signal stimulus and $E4$ receiving the random stimulus changed little. Thus, these networks gradually learned to predict the temporal sequence, while the random stimulus was not predicted. This implies that the inhibitory neuron suppressed spiking of neurons  $E1$-$3$ at the time of target input and that small networks with synaptic delay can learn to predict temporal sequences.

\subsection{Large Random Networks}
\begin{figure}[htbp]
\begin{center}
\includegraphics[width=9cm]{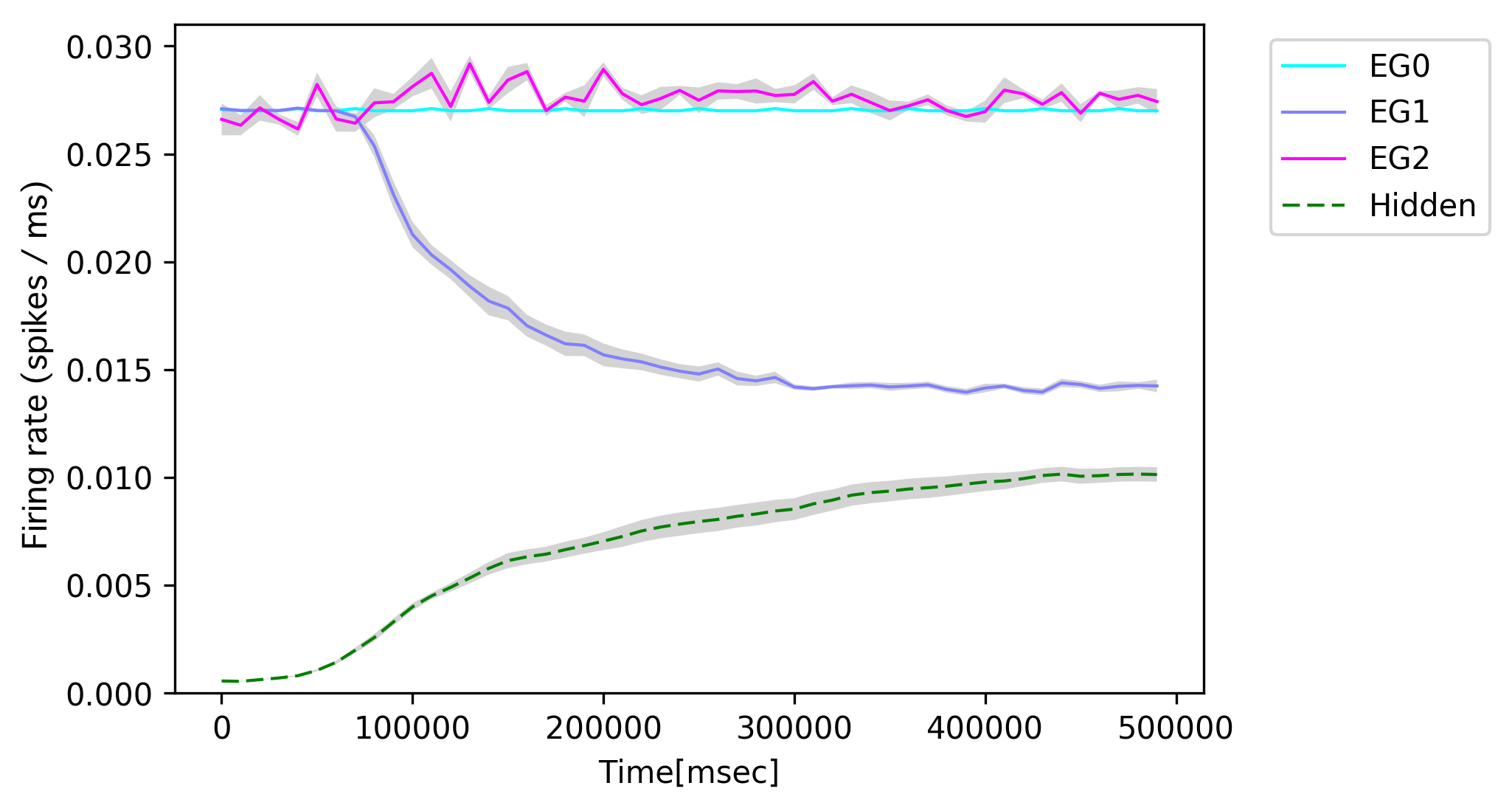}
\end{center}
\caption[Time series of the firing rates of the input neurons with the minimal pattern in the large networks (100 neurons).]{Time series of the firing rates of the input neurons with the minimal pattern in the large networks (100 neurons). The shaded regions represent the standard errors of the mean (20 networks). The firing rate of $EG1$ which got the target stimuli decreased but $EG0$ which got the signal stimuli and $E2$ which got the random stimuli did not change much. The firing rate of hidden neurons can be regarded as a baseline of the firing rate in the network.}
\label{fig:prediction_randomnet_firingrate}
\end{figure}

We then examined whether the learning properties of these minimal networks are scalable to large networks. We applied the minimal pattern of stimulation to random networks with 100 neurons. Figure~\ref{fig:prediction_randomnet_firingrate} shows that these larger networks gradually learned to predict the minimal pattern, while the random stimulation was not predicted. The green line represents the firing rate of hidden neurons, which can be regarded as the baseline firing rate of the network. The firing rate of excitatory neuron group 1 ($EG1$) receiving target stimuli gradually decreased to near baseline levels, indicating reasonable prediction accuracy.

\begin{figure}[htbp]
\begin{center}
\includegraphics[width=9cm]{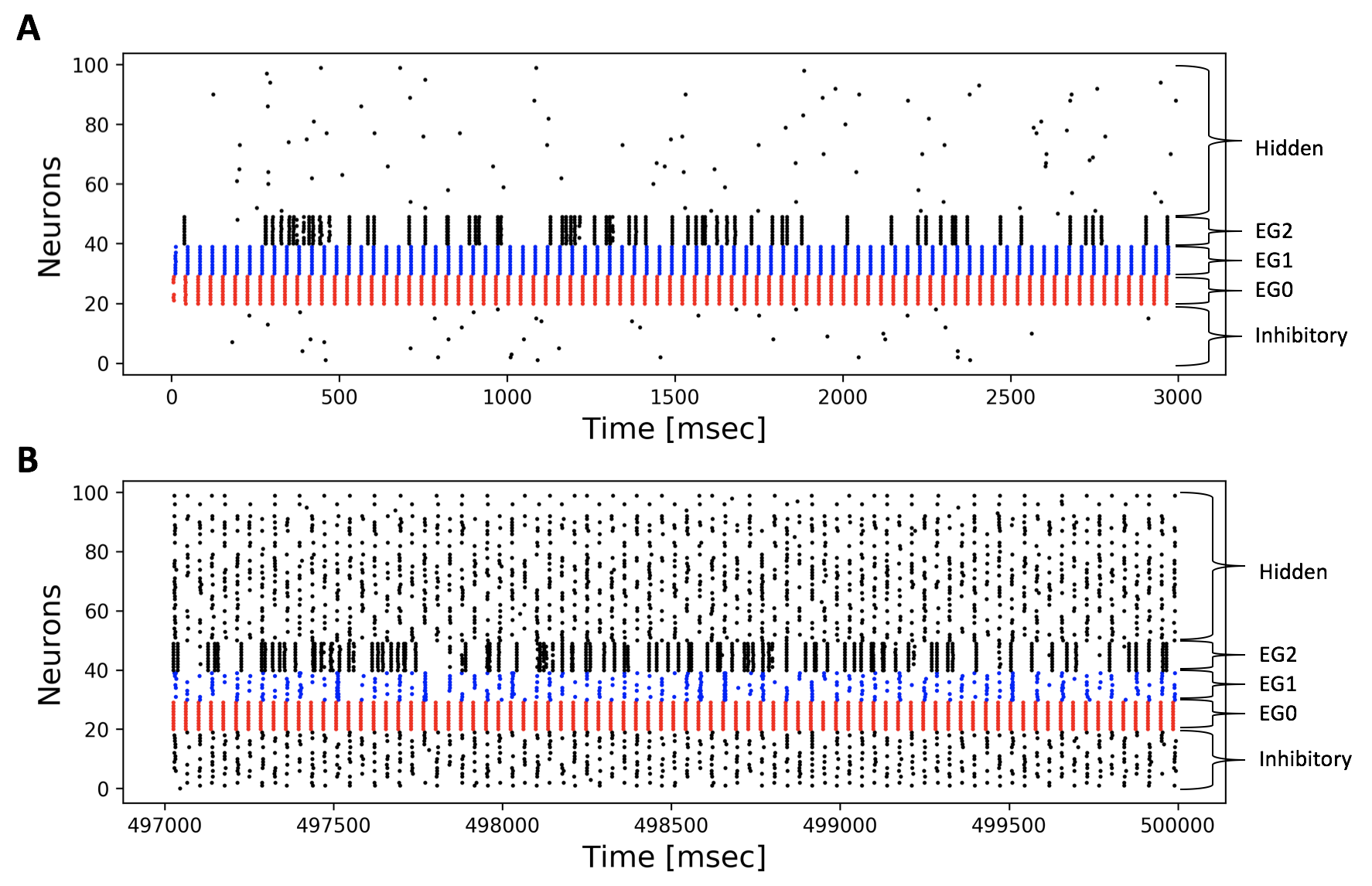}
\end{center}
\caption[Raster plots of spikes in large network.]{Raster plots of spikes in large networks. Each dot represents one spike: the red dots represent spikes of $EG0$ which got the signal stimuli, the blue dots represent spikes of $EG1$ which got the target stimuli, black dots represent spikes of the other groups ($EG2$, Hidden and Inhibitory). A: Spikes in the first 3,000 ms. Almost all neurons in $EG1$ fired at the timing they got the target stimuli. B: Spikes in the last 3,000 ms. The neurons in EG1 did not fire much at the timing of the stimulation and their firing patterns were almost the same as the hidden neurons. }
\label{fig:prediction_randomnet_raster}
\end{figure}

Figure~\ref{fig:prediction_randomnet_raster} shows a typical example of raster plot of spikes for each neuron in the large network. In the first phase of the experiment (first 3000 ms), almost all neurons of group $EG1$ fired immediately in response to the target stimuli while in the later phase (final 3000 ms), the firing rate was much lower and the pattern was very similar to that of the hidden neurons. This implies that inhibitory neurons suppressed the firing of $EG1$ neurons at the time of target stimulation, suggesting that large random networks can learn to predict the minimal pattern of stimulation using only STDP. 

\begin{figure}[htbp]
\begin{center}
\includegraphics[width=9cm]{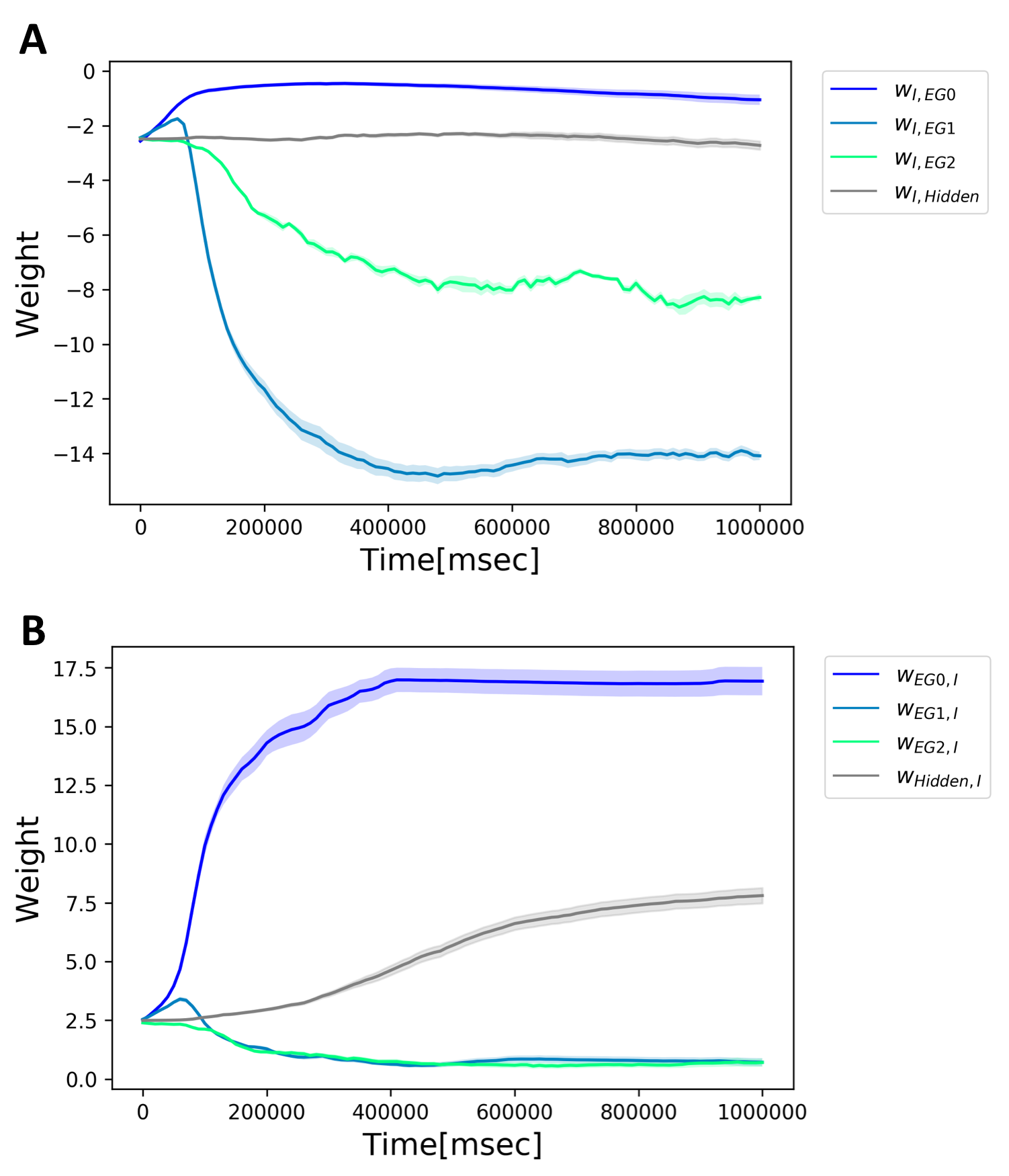}
\end{center}
\caption[Time series of synaptic weights in large networks with minimal pattern.]{Time series of synaptic weight changes in large networks during minimal pattern stimulation. A: Weights of connections from the inhibitory neurons to the excitatory neuron groups $EG0$, $EG1$, $EG2$, and hidden neurons. B: Weight from the excitatory neuron groups $EG0$, $EG1$, $EG2$, and hidden neurons to the inhibitory neurons. }
\label{fig:prediction_randomnet_weight_series}
\end{figure}

Moreover, the time series of synaptic weight changes between neuron groups resembles those of the small networks (Fig.~\ref{fig:prediction_randomnet_weight_series}), suggesting that the mechanisms underlying prediction are similar.
 
\begin{figure}[htbp]
\begin{center}
\includegraphics[width=7cm]{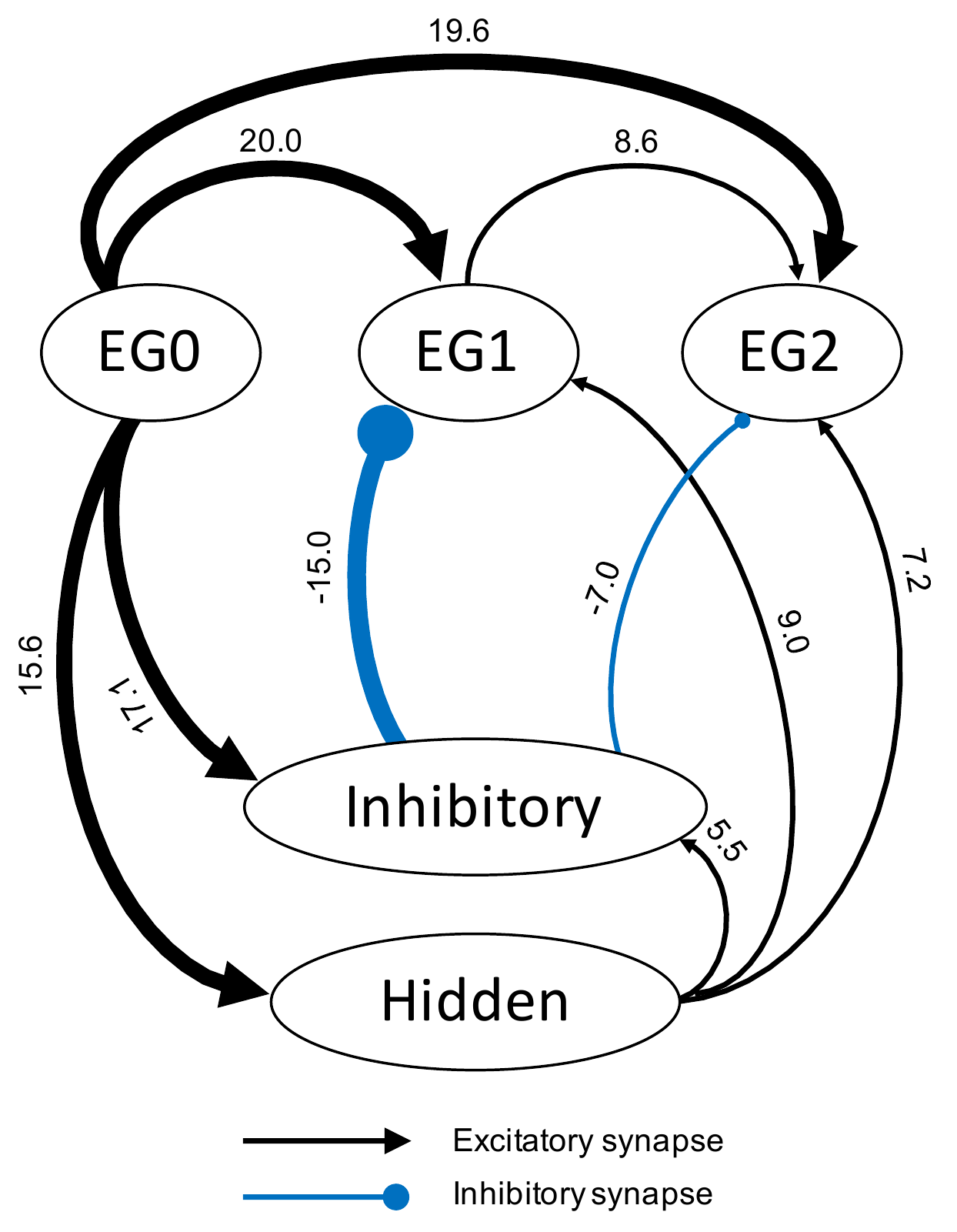}
\end{center}
\caption[Final topology of the large network.]{Final topology of the large network resulting from STDP. A connection between each group is depicted if the absolute mean weight value is larger than 5.0. The thick lines represent weight values larger than 10.0 and thin lines weight values lower than 10.0. The black arrows represent excitatory synapses and the blue lines inhibitory synapses. The weight values of connections from inhibitory neurons are negative.}
\label{fig:prediction_randomnet_topology}
\end{figure}

Figure~\ref{fig:prediction_randomnet_topology} shows the final topology of the network. There was a strong pathway from $EG0$ to inhibitory neurons and from inhibitory neurons to $EG1$; This pathway is required for the prediction. In addition, there was a pathway from $EG0$ to the hidden neurons, from the hidden neurons to the inhibitory neurons, and from the inhibitory neurons to $EG1$. This pathway may be required to adjust the timing of $EG1$ suppression.


Collectively, these findings demonstrate that the learning performance of small minimal networks is scalable to larger random networks. Thus, spiking neural networks without specific structure (random networks) can spontaneously learn to predict simple stimulus sequences 
based solely on STDP.

\section{Discussion}
We demonstrate that spiking neural networks can spontaneously learn to predict input sequences from the environment based only on STDP. 
The firing of input neurons receiving target stimuli (those following signaling inputs) were predicted and suppressed by inhibitory neurons, thereby reducing the influence of environmental stimulation on network activity. In other words, neural networks can learn to avoid (ignore) specific input stimulation pattern through STDP. 

We also demonstrate that this property is scalable from small networks of a few neurons to larger random networks of 100 neurons without specific design structures or other functions except STDP. These findings also suggest that like stimulation avoidance by action and selection in neuronal cultures \cite{Masumori2015, Masumori2018}, prediction can emerge to avoid stimulation in neuronal cultures via STDP, although this notion remains to be confirmed.

Our previous studies showed that action and selection can emerge in spiking neural networks based on STDP and that these mechanisms can eliminate specific stimulus inputs \cite{Masumori2019}. In this study, we found that stimulation avoidance by prediction also emerges in spiking neural networks under similar conditions. This findings implies that neural networks can establish at least three mechanisms to eliminate (avoid) specific stimulation patterns based on STDP: action, prediction, and selection. We also speculate that these three mechanisms can emerge in the same neural network with STDP depending on the quality of the stimuli. Specifically, controllable inputs induce action, predictable inputs induce prediction (there might be controllable and predictable input, and such input induce action or prediction) and uncontrollable inputs (noise) induce selection. In other words, distinct mechanisms can emerge for stimulus avoidance under specific environmental conditions. We call this the principle of stimulus avoidance (PSA). In future work, we will test whether action, prediction and selection actually emerge in same network {\it in silico} and {\it in vitro} according to the PSA depending on differences in stimulus input properties. 

\section*{Acknowledgment}
This work is partially supported by MEXT project ``Studying a Brain Model based on Self-Simulation and Homeostasis'' in Grant-in-Aid for Scientific Research on Innovative Areas ``Correspondence and Fusion of Artificial Intelligence and Brain Science'' (19H04979)



%



\bibliographystyle{IEEEtran}
\bibliography{99_C_Phd_Thesis_masumori_2.bib}

\begin{thebibliography}{10}
\providecommand{\url}[1]{#1}
\csname url@samestyle\endcsname
\providecommand{\newblock}{\relax}
\providecommand{\bibinfo}[2]{#2}
\providecommand{\BIBentrySTDinterwordspacing}{\spaceskip=0pt\relax}
\providecommand{\BIBentryALTinterwordstretchfactor}{4}
\providecommand{\BIBentryALTinterwordspacing}{\spaceskip=\fontdimen2\font plus
\BIBentryALTinterwordstretchfactor\fontdimen3\font minus
  \fontdimen4\font\relax}
\providecommand{\BIBforeignlanguage}[2]{{%
\expandafter\ifx\csname l@#1\endcsname\relax
\typeout{** WARNING: IEEEtran.bst: No hyphenation pattern has been}%
\typeout{** loaded for the language `#1'. Using the pattern for}%
\typeout{** the default language instead.}%
\else
\language=\csname l@#1\endcsname
\fi
#2}}
\providecommand{\BIBdecl}{\relax}
\BIBdecl

\bibitem{Bastos2012}
A.~M. Bastos, W.~M. Usrey, R.~A. Adams, G.~R. Mangun, P.~Fries, and K.~J.
  Friston, ``{Canonical Microcircuits for Predictive Coding},'' pp. 695--711,
  2012.

\bibitem{Clark2013}
A.~Clark, ``Whatever next? predictive brains, situated agents, and the future
  of cognitive science,'' \emph{Behavioral and Brain Sciences}, vol.~36, no.~3,
  pp. 181--204, 2013.

\bibitem{Rao1999}
R.~P.~N. Rao and D.~H. Ballard, ``{Predictive coding in the visual cortex: A
  functional interpretation of some extra-classical receptive-field effects},''
  \emph{Nature Neuroscience}, vol.~2, no.~1, pp. 79--87, 1999.

\bibitem{Huang2011}
Y.~Huang and R.~P.~N. Rao, ``Predictive coding,'' \emph{Wiley Interdisciplinary
  Reviews: Cognitive Science}, vol.~2, no.~5, pp. 580--593, 2011.

\bibitem{Friston2009}
K.~Friston and S.~Kiebel, ``Predictive coding under the free-energy
  principle,'' \emph{Philosophical Transactions of the Royal Society B:
  Biological Sciences}, vol. 364, no. 1521, pp. 1211--1221, 2009.

\bibitem{Lotter2017}
W.~Lotter, G.~Kreiman, and D.~Cox, ``Deep predictive coding networks for video
  prediction and unsupervised learning,'' in \emph{5th International Conference
  on Learning Representations, {ICLR} 2017, Toulon, France, April 24-26, 2017,
  Conference Track Proceedings}, 2017.

\bibitem{Song2000}
S.~Song, K.~D. Miller, and L.~F. Abbott, ``{Competitive Hebbian learning
  through spike-timing-dependent synaptic plasticity.}'' \emph{Nature
  neuroscience}, vol.~3, no.~9, pp. 919--26, sep 2000.

\bibitem{Sinapayen2016}
L.~Sinapayen, A.~Masumori, and T.~Ikegami, ``{Learning by stimulation
  avoidance: A principle to control spiking neural networks dynamics},''
  \emph{PLOS ONE}, vol.~12, no.~2, p. e0170388, feb 2017.

\bibitem{Masumori2017a}
A.~Masumori, L.~Sinapayen, and T.~Ikegami, ``{Learning by stimulation avoidance
  scales to large neural networks},'' in \emph{Proceedings of the 14th European
  Conference on Artificial Life ECAL 2017}.\hskip 1em plus 0.5em minus
  0.4em\relax Cambridge, MA: MIT Press, sep 2017, pp. 275--282.

\bibitem{Masumori2017b}
A.~Masumori, N.~Maruyama, L.~Sinapayen, T.~Mita, U.~Frey, D.~Bakkum,
  H.~Takahashi, and T.~Ikegami, ``{Learning by Stimulation Avoidance in
  Cultured Neuronal Cells},'' in \emph{The 2nd International Symposium on Swarm
  Behavior and Bio-Inspired Robotics (SWARM2017)}, 2017.

\bibitem{Masumori2018}
A.~Masumori, L.~Sinapayen, N.~Maruyama, T.~Mita, D.~Bakkum, U.~Frey,
  H.~Takahashi, and T.~Ikegami, ``Autonomous regulation of self and non-self by
  stimulation avoidance in embodied neural networks,'' \emph{The 2018
  Conference on Artificial Life: A Hybrid of the European Conference on
  Artificial Life (ECAL) and the International Conference on the Synthesis and
  Simulation of Living Systems (ALIFE)}, pp. 163--170, 2018.

\bibitem{Buonomano2000}
D.~V. Buonomano, ``Decoding temporal information: A model based on short-term
  synaptic plasticity,'' \emph{Journal of Neuroscience}, vol.~20, no.~3, pp.
  1129--1141, 2000.

\bibitem{Rao2001}
R.~P.~N. Rao and T.~J. Sejnowski, ``Spike-timing-dependent hebbian plasticity
  as temporal difference learning,'' \emph{Neural Comput.}, vol.~13, no.~10,
  pp. 2221--2237, Oct. 2001.

\bibitem{Wacongne2012}
C.~Wacongne, J.-P. Changeux, and S.~Dehaene, ``{A Neuronal Model of Predictive
  Coding Accounting for the Mismatch Negativity},'' \emph{Journal of
  Neuroscience}, vol.~32, no.~11, pp. 3665--3678, 2012.

\bibitem{Izhikevich2003a}
E.~Izhikevich, ``{Simple model of spiking neurons},'' \emph{IEEE Transactions
  on Neural Networks}, vol.~14, no.~6, pp. 1569--1572, 2003.

\bibitem{Masumori2015}
A.~Masumori, N.~Maruyama, L.~Sinapayen, T.~Mita, U.~Frey, D.~Bakkum,
  H.~Takahashi, and T.~Ikegami, ``{Emergence of Sense-Making Behavior by the
  Stimulus Avoidance Principle: Experiments on a Robot Behavior Controlled by
  Cultured Neuronal Cells},'' \emph{Proc. of the European Conference on
  Artificial Life (ECAL) 2015}, pp. 373--380, 2015.

\bibitem{Masumori2019}
A.~Masumori, L.~Sinapayen, N.~Maruyama, T.~Mita, D.~Bakkum, U.~Frey,
  H.~Takahashi, and T.~Ikegami, ``{Neural Autopoiesis: Organizing Self-Boundary
  by Stimulus Avoidance in Biological and Artificial Neural Networks},''
  \emph{Artificial Life}, vol. in press, 2019.

\end{thebibliography}

\end{document}